%% file: main.tex
\documentclass[11pt]{article} %
\usepackage{aaai22}
\usepackage{times}  %
\usepackage{helvet}  %
\usepackage{courier}  %
\usepackage{graphicx} %
\usepackage{natbib}  %
\usepackage{caption} %
\DeclareCaptionStyle{ruled}{labelfont=normalfont,labelsep=colon,strut=off} %
\usepackage{algorithm}
\usepackage{algorithmic}

\usepackage{latexsym}
\usepackage{booktabs}
\usepackage{amssymb}
\usepackage{xspace}
\usepackage{multirow}
\usepackage{fancyvrb}
\graphicspath{ {./figs/} }
\usepackage[T1]{fontenc}

\usepackage[utf8]{inputenc}

\usepackage{amsmath}
\usepackage{bm}
\usepackage{listings}
\usepackage{tablefootnote}
\usepackage{microtype}
\usepackage{xcolor}
\usepackage[capitalise]{cleveref}

\usepackage{color}
\definecolor{deepblue}{rgb}{0,0,0.5}
\definecolor{deepred}{rgb}{0.6,0,0}
\definecolor{deepgreen}{rgb}{0,0.5,0}
\definecolor{darkgreen}{RGB}{43,163,39}
\definecolor{bluesquare}{rgb}{126,166,224}
\definecolor{LightGray}{gray}{0.9}
\definecolor{DarkGray}{gray}{0.1}

\renewcommand{\tt}[1]{\fontfamily{cmtt}\selectfont #1}

\lstdefinestyle{pythoncode}{
	language=Python,
	otherkeywords={self,join,append,split,write},   %
	keywordstyle=\bfseries\color{deepblue},
	emph={__init__},          %
	emphstyle=\color{deepred},    %
	showstringspaces=false,
	breaklines=true,
	escapeinside=||,
	columns=fullflexible,
	basicstyle=\fontfamily{cmtt}\scriptsize,
    belowskip=-\baselineskip,
    aboveskip=-0.7\baselineskip
}

\definecolor{codegreen}{rgb}{0,0.6,0}
\definecolor{codegray}{rgb}{0.5,0.5,0.5}
\definecolor{codepurple}{rgb}{0.58,0,0.82}
\definecolor{backcolour}{rgb}{0.95,0.95,0.92}

\newcommand{\aarg}[3]{\textcolor{blue}{\tt{#1}} \textcolor{black}{\tt{#2}} \textcolor{black}{\tt{#3}}}
\newcommand{\aargt}[2]{\textcolor{blue}{\tt{#1}} \textcolor{black}{\tt{#2}}}

\newcommand{\actiont}[1]{\textcolor{blue}{\tt{#1}}}
\newcommand{\objectt}[1]{\textcolor{black}{\tt{#1}}}

\newcommand\xv{\ensuremath{\bm{x}}\xspace}
\newcommand\E{\ensuremath{\mathcal{E}}\xspace}
\newcommand\Aa{\ensuremath{\mathcal{A}^a}\xspace}
\newcommand\Ap{\ensuremath{\mathcal{A}^p}\xspace}
\newcommand\A{\ensuremath{\mathcal{A}}\xspace}
\newcommand\av{\ensuremath{\bm{a}}\xspace}
\newcommand\rv{\ensuremath{\bm{r}}\xspace}
\newcommand{\eg}{\hbox{\emph{e.g.}}\xspace}
\newcommand{\ie}{\hbox{\emph{i.e.}}\xspace}
\newcommand{\newcolor}{black}

\usepackage{newfloat}
\usepackage{listings}
\floatstyle{ruled}
\newfloat{listing}{tb}{lst}{}
\floatname{listing}{Listing}
\title{Procedures as Programs: Hierarchical Control of Situated Agents \\ through Natural Language}
\author{
 Shuyan Zhou, Pengcheng Yin, Graham Neubig \\
 Language Technologies Institute \\
 Carnegie Mellon University \\
  \texttt{\{shuyanzh, pcyin, gneubig\}@cs.cmu.edu}\ \\
}

\usepackage{bibentry}

\begin{document}

\maketitle

\begin{abstract}
When humans conceive how to perform a particular task, they do so hierarchically: splitting higher-level tasks into smaller sub-tasks.
However, in the literature on natural language (NL) command of situated agents, most works have treated the procedures to be executed as flat sequences of simple actions, or any hierarchies of procedures have been shallow at best.
In this paper, we propose a formalism of \emph{procedures as programs}, a powerful yet intuitive method of representing hierarchical procedural knowledge for agent command and control.
We further propose a modeling paradigm of \emph{hierarchical modular networks}, which consist of a \emph{planner} and \emph{reactors} that convert NL intents to predictions of executable programs and probe the environment for information necessary to complete the program execution.
We instantiate this framework on the IQA and ALFRED datasets for NL instruction following. Our model outperforms reactive baselines by a large margin on both datasets.
We also demonstrate that our framework is more data-efficient, and that it allows for fast iterative development.%
\end{abstract}

\section{Introduction}\label{intro}

Procedural knowledge, or ``how-to'' knowledge, refers to knowledge about the execution of particular tasks.
It is inherently hierarchical; high-level procedures consist of many lower-level procedures. For example, ``cooking a pizza'' comprises many lower-level procedures, including ``buying ingredients'', ``pre-heating the oven'', etc. There are also multiple levels of hierarchy; ``buying ingredients'' can be further decomposed to ``going to the grocery'', ``paying'' etc.

\begin{figure*}[t!]
    \centering
    \includegraphics[width=\textwidth]{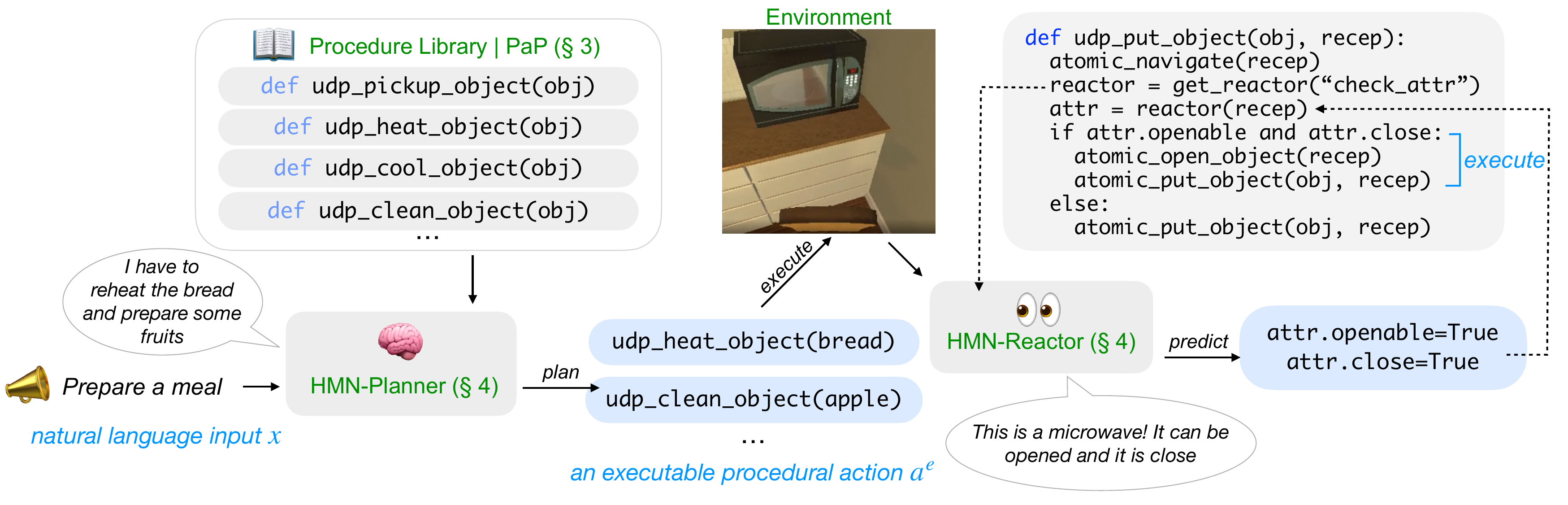}
    
    \caption{The proposed framework, containing a hierarchical library of procedures written as Python functions (\S\ref{sec:pap}). Coupled with this library is a hierarchical neural network (HMN, \S\ref{sec:hmn}) with a \textsc{Planner} that constructs an executable procedure and \textsc{Reactors} that react to the environment to resolve control flow.}
    \label{fig:framework}
\end{figure*}

There has been significant prior work on benchmarks and methods for complex task completion using situated agents given natural language (NL) instructions, such as agents trained to navigate the web and mobile UIs~\cite{li2020mapping, xu2021grounding} or solve household tasks \cite{shridhar_alfred:_2019}.
However, the great majority of methods used to solve these tasks use a \emph{reactive} strategy that makes decisions on the lowest-level atomic actions available to the agent while making steps through the environment \cite{gupta2017cognitive, zhu2020vision}. In works that attempt to define procedures more explicitly, they are often defined in a shallow way where there only exists one level of hierarchy~\cite{andreas_modular_2017, gordon2018iqa, yu2019multi, das_neural_2019}.%
~Besides, these methods are less data-efficient. NL is abstract, and a deceptively simple instruction contains multiple unspoken procedures that the human users assume that their audiences would be able to reason about. Due to this semantic gap, most existing works require a significant amount of labeled data or trial-and-error to learn to map NL to atomic actions. The performance is also limited on long-horizon tasks due to compounding errors~\cite{xu2019regression}. \textcolor{\newcolor}{Finally, these works generally define policy sketches through natural language utterances, which are not directly executable; they only serve as additional conditioning that informs the generation of actions.} 

In this work, we propose a \emph{framework for representing and leveraging hierarchical procedural knowledge} to improve the control of situated agents accomplishing complex tasks described in natural language (example in \cref{fig:framework}). We first propose a method for representing procedures as programs (PaP) written in a high-level programming language like Python (\S\ref{sec:pap}). \textcolor{\newcolor}{The execution of each program yields actions to accomplish a task described in NL.}
There are several merits to this formalism.
First, programs are inherently hierarchical; they apply nested function calls to realize higher-level functionality with multiple calls to lower-level functionality.
Second, programs have built-in control-flow operators, enabling a general representation to deal with multiple divergent situations without the loss of higher-level abstraction. 
Third, programs provide a flexible way to define, share and call different machine-learned components to perceive the environment through an embodied agent's executions. These three features remain largely unexplored in the existing representations~\cite{chen2011learning, artzi2013weakly,misra2016tell}, as discussed further in \S\ref{sec:contrast}.
Finally, programs in a high-level language are comprehensible and curatable, allowing for fast development on various tasks.

Coupled with this representation, we propose a modeling paradigm of \emph{hierarchical modular networks} (HMN; \S\ref{sec:hmn}) that has (1) a \textsc{Planner} that maps the NL instructions to their corresponding executable programs and (2) a collection of \textsc{Reactors} that perceive the environment and provide context-sensitive feedback to decide the further execution of the program. Such modular design can facilitate training efficiency as well as improve the performance of each individual component~\cite{andreas_neural_2016}. 

We instantiate our framework on two task settings: the IQA dataset~\cite{gordon2018iqa} where an agent explores the environment to answer questions regarding objects; and the ALFRED dataset \cite{shridhar_alfred:_2019}, in which an agent must map natural language instructions to actions to complete household tasks (\S\ref{sec:instance}).
In experiments (\S\ref{sec:experiments}), we find that our framework outperforms the reactive baseline by a significant margin on both datasets, and is significantly more data-efficient. We also demonstrate the flexibility of our framework for fast iterative development of program libraries. \textcolor{\newcolor}{We end with a discussion of the limitations of the framework and the potential solutions, paving the way for future works that scale our framework to more open-domain tasks (\S7)}.

\input{1_problem_formulation}

\input{2_instance}
\input{3_experiments}
\input{5_discussion}
\bibliography{aaai22}
\bibliographystyle{acl_natbib}
\clearpage
\appendix
\input{6_appendix}
\end{document}

%% file: 1_problem_formulation.tex
\section{Task: Controlling Situated Agents}
\label{definitions}

First, we define the task of controlling an agent
in some situated environment \E~through natural language.
The environment \E~provides a set of atomic actions $\Aa=\{a^a_1, a^a_2, ..., a^a_h\}$ to interact with the environment.
Each atomic action can take zero or more arguments that specify which parts of the environment to which it is to be applied.
We denote action $a^a_i$'s $j$th argument as $r_{i,j}$.
The specific type of each argument will depend on the action and environment \E; it could be discrete symbols, scalar values, tensors describing regions of the visual space with which to interact, etc. \textcolor{\newcolor}{The environment also defines the transition of the state after executing an atomic action.}
Given a user intent \xv described in natural language, the control system aims at creating an atomic action sequence consisting of a sequence of actions $\av = [a_1, a_2, ..., a_n]$ ($a_i \in \Aa$) and concrete assignments \rv of the action arguments for each of these $n$ actions.
This action sequence is executed against the environment to achieve a result $\hat{y} = \E(\av,\rv)$, which is compared against a gold-standard result $y$ using a score function $s(y,\hat{y})$. Action sequences realizing the intent behind $\xv$ will receive a high score, and those that do not will receive a low score.

\begin{table}[t]
  \centering
  \footnotesize
  \renewcommand{\tabcolsep}{2pt}
  \begin{tabular}{rp{15cm}}
  \toprule
\begin{lstlisting}[numbers=none, basicstyle=\fontfamily{cmtt}\small,style=pythoncode,belowskip=-\baselineskip,aboveskip=- 0.5\baselineskip,commentstyle=\color{codegreen}]
# C1: an atomic action to toggle on an appliance
def atomic_toggle_on(obj):
    env.call("toggle_on", obj)
# C2: a procedural action to pick and then put an object
def udp_pick_and_put_object(obj, dst):
    udp_pickup_object(obj)
    udp_put_object(obj, dst)
# C3: an emptying receptacle procedure with for-loop 
def udp_empty_recep(recep, dst):
    reactor = get_reactor("find_all_obj")
    obj_list = reactor(recep)
    for obj in obj_list:
        udp_pick_and_put_object(obj, dst)
# C4: a pickup object procedure with control flow
def udp_pickup_object(obj):
    atomic_navigate(obj)
    reactor1 = get_reactor("find_recep")
    reactor2 = get_reactor("check_obj_attr")
    recep = reactor1(obj)
    attr = reactor2(recep)
    if attr.openable and attr.close:
        atomic_open_object(recep)
        atomic_pickup_object(obj)
        atomic_close_object(recep)
    else: atomic_pickup_object(obj)
 
\end{lstlisting}
\\ \bottomrule
\end{tabular}
\caption{Action functions written in Python. Atomic action function starts with atomic and a procedural action function starts with {\tt udp}.}
\label{code-cases}
\end{table}        

\section{Representing Procedures as Programs}
\label{sec:pap}

\label{sec:papactions}

Our PaP formalism consists of two types of actions:
\emph{atomic actions} that can be issued directly to the environment, and \emph{procedural actions} that abstractly describe the higher-level procedures.
Since both action types are implemented as functions, we use ``action'' and ``function'' interchangeably. A few examples are listed in \cref{code-cases}.

\paragraph{Interface to Atomic Actions \Aa (C1)}~Atomic actions provide a medium for direct interaction with the environment. 
The call of an atomic action with proper argument types will invoke the corresponding execution in the environment. 

\paragraph{Procedural Actions \Ap (C2-C4)}~Procedural actions describe abstractions of higher-level procedures composed of either lower-level procedures or atomic actions. Importantly, lower-level procedures can be \emph{re-used} across multiple higher-level procedures without re-definition. Formalizing the hierarchies in this compact way can not only facilitate the procedure library curation process but also potentially benefit automatic procedure library induction (e.g.~through minimal description length~\cite{ellis2020dreamcoder}).

\paragraph{Control-flow of \Ap (C3-C4)}~There can be multiple execution traces to accomplish the same goal under different conditions. For example, picking up an object from inside a closed receptacle requires opening the receptacle first, while the open action is not required for objects not in a receptacle. 
To improve the \emph{coverage} of procedural functions we leverage the built-in control flow  of the host programming language to allow for conditional execution of environment-specific actions. 
In the body of a procedural function, we can use control flow to define divergent branches to handle different situations (\textbf{C4}). To deal with the repeated calls of the same routine, we further introduce for/while-loops. For example, \textbf{C3} works for emptying receptacles with variable number of objects without repeatedly writing down the {\tt udp\_pick\_put\_object}. Leveraging control flows to describe divergent procedural traces remains largely unexplored in previous works. 

\paragraph{Call of Situated Components (\textbf{C3}-\textbf{C4})}~We define control flow that can be dynamically triggered upon different states of the environment, which often remain unknown before the agent interacts with the environment. 
We introduce situated components to probe the environment and gather state information to guide program execution.
In \textbf{C4}, the agent uses two different reactors to find the potential holder of an object ({\tt reactor1}) and exam the holder's properties ({\tt reactor2}). A reactor can be implemented in many ways~(\eg~using a neural network). 

\subsection{Contrast to Previous Formalisms}
\label{sec:contrast}
In contrast to most previous works that employ domain-specific formalisms like lambda calculus to represent procedural knowledge~\cite{artzi2013weakly, artzi-das-petrov:2014:EMNLP2014}, PaP uses widely-adopted general-purpose programming languages (\eg Python) to specify and represent hierarchical procedures.
These are more comprehensible and do not require system designers to learn a new task-specific language.
They also enable easy creation of more hierarchical procedures with \emph{reusable sub-routines}. Existing works~\cite{chen_compositional_2020,artzi2013weakly,misra2016tell, das_neural_2019} do not model such sub-procedures as independent components, and simply define procedures as a flat sequence of actions without any hierarchy. 
Our approach of representing hierarchical procedures using reusable sub-routines is also reminiscent of recent works in semantic parsing, which compose complex programs from learned idiomatic program structures~\cite{Raghothaman2016SWIMSW, iyer_learning_2017, shin_program_2019}. Readers are referred to \S\ref{sec:related_work} for more discussion.

Additionally, PaP uses control flow with divergent branches to handle environment-specific variations of a high-level procedure.
A single procedure could therefore dynamically adapt to a variety of environments following the branches triggered by the environments.
This makes our representations more compact.
To our best knowledge, this feature is largely unexplored in the literature.

Finally, PaP provides a convenient interface for procedures to query and interact with task-specific situated components.
Under PaP, situated components are exposed as pre-defined APIs, and could be easily called by high-level procedures.  
In contrast, existing works either require separate mechanisms to call probing components~\cite{misra2016tell}, or their working environment is less complicated, and the flexible use of a collection of situated components is not a necessity~\cite{chen2011learning}.  

We can also view the PaP formalism as a way to construct behavior trees~\cite{colledanchise2018behavior} which have been used in robotic planning and game design literature.
We can use the off-the-shelf tools to convert the programs to abstract syntax trees (AST) which resemble these trees.

\section{Hierarchical Modular Networks}
\label{sec:hmn}
This section introduces how to use the procedure library \A to generate executable programs to complete tasks described in natural language \xv. We propose a modeling method of \emph{hierarchical modular networks} (HMN) that consists of two main components. First, there is a \textsc{HMN-Planner} that convert \xv to an executable procedural action $\av^{e}=\{a_1, a_2, ..., a_n\}$ where $a_i$ either belongs to atomic functions \Aa or procedural functions \Ap. We model the \textsc{HMN-Planner} as a sequence-to-sequence model where the encoder takes \xv as input, and the decoder generates one function $a_i$ at a time from a constrained vocabulary $\Ap \bigcup \Aa$, conditioned on \xv and the action history $\{a_1, ..., a_{i-1}\}$. %

Next, we define the collection of situated components, ``reactors,'' as \textsc{HMN-Reactors}. Each reactor is a classifier that predicts one or many labels given the observed information (\eg the NL input, the visual observation. %
For example, {\tt reactor2} in \textbf{C4} in \cref{code-cases} probes the status of a receptacle based on receptacle name and the visual input. \textsc{HMN-Reactors} allows us to flexibly share the same reactor among different functions and design separated reactors to serve different purposes. For example in \textbf{C4}, we use two reactors to find the possible receptacle of an object ({\tt reactor1}) and to perceive the open/closed status of a receptacle ({\tt reactor2}) since these two tasks presumably require more mutually exclusive information. %
At the same time, we share {\tt reactor2} to also probe the related {\tt openable} property of a receptacle for more efficient parameter sharing. This sort of modular design leads to efficient training and improved performance~\cite{andreas_neural_2016}.

%% file: 2_instance.tex
\section{Instantiations}\label{sec:instance}
In this section, we introduce two concrete realizations of the proposed framework over the IQA dataset \cite{gordon2018iqa} and the ALFRED dataset \cite{shridhar_alfred:_2019}. %
Both are based on egocentric vision in a high-fidelity simulated environment THOR \cite{robothor}.

\subsection{IQA}\label{sec:iqa}
IQA is a dataset for situated question answering with three types of questions querying (1) the existence of an object (\eg \textit{Is there a mug?}), (2) the count of an object (\eg \textit{How many mugs are there?}) and (3) whether a receptacle contains an object (\eg \textit{Is there a mug in the fridge?}). 

There are seven atomic actions in IQA, \ie \actiont{Moveahead}, \actiont{RotateLeft}, \actiont{RotateRight}, \actiont{LookDown}, \actiont{LookUp}, \actiont{Open} and \actiont{Close}; and all arguments are expressed through the unique object IDs (\eg \objectt{apple\_1}). We further process the atomic navigation actions to a single atomic action \actiont{Navigate} with one argument \objectt{destination}, which moves the agent directly to the destination. This replacement is done by searching the scene and recording the coordinates of unmovable objects (\eg cabinet) -- more details provided in the Appendix~\ref{sec:pre_search}.

\paragraph{Procedure Library}~We design a procedure for each of the three types of questions in IQA, as shown in \cref{iqa-procedure}.
Generally speaking, those procedures first search all or a subset of the receptacles (\eg {\tt table}, {\tt fridge}) in a scene for the target object (\eg {\tt mug}), and then execute a question-specific intent (\eg existence-checking, counting).
\cref{iqa-procedure} shows the procedure for answering existence questions.
Since the target object can be inside a receptacle (\eg fridge), we introduce control flow to decide whether to open and close a receptacle before and after checking its contents in sub-procedure {\tt udp\_check\_relation}. Following the paper author's understanding of the three types of questions, these procedural functions were created without looking into any actual trajectories that answer these questions. 
In total, we define six procedural actions with a complete list in Appendix~\ref{app:full_library}.

\begin{table}[t]
  \centering
  \footnotesize
  \renewcommand{\tabcolsep}{2pt}
  \begin{tabular}{rp{15cm}}
  \toprule
\begin{lstlisting}[numbers=none, basicstyle=\fontfamily{cmtt}\small,style=pythoncode,belowskip=-\baselineskip,aboveskip=- 0.5\baselineskip,commentstyle=\color{codegreen}]
# check existence of an object in the scene
def udp_check_obj_exist(obj):
    all_recep = udp_grid_search_recep()
    for recep in all_recep:
        rel = udp_check_relation(obj, recep)
        if rel == OBJ_IN_RECEP:
            return True
    return False
# check object inside receptacle
def udp_check_relation(obj, recep):
    atomic_navigate(recep)
    r1 = get_reactor("check_obj_attr")
    r2 = get_reactor("check_obj_recep_rel")
    attr = r1(recep)
    if attr.is_openable and attr.is_closed:
        atomic_open_object(recep)
        rel = r2(obj, recep)
        atomic_close_object(recep)
    else:
        rel = r2(obj, recep)
    return rel
\end{lstlisting}
\\ \bottomrule
\end{tabular}
\caption{The procedural actions to answer the existence questions of the IQA dataset.}
\label{iqa-procedure}
\end{table}

\paragraph{HMN}~The natural language questions \xv in IQA are generated with a limited number of templates. There are only seven receptacles, and three of them are openable. We thus use a rule-based \textsc{HMN-Planner} to map a template to one of the three high-level procedural actions (\ie existence, count and contain). Then, we design two reactors, each as a multi-classes classifier: \textsc{AttrChecker}, %
which examines the properties (whether the object is openable) and the status (whether the object is opened) of an object, and \textsc{RelChecker}, which checks the spatial relation between two objects. We leave the detailed implementations of the reactors to Appendix~\ref{sec:iqa_hmn}. Notably, we use \emph{zero} IQA training data to build the HMN. Instead, it is made up of a few heuristic components based on the predictions of a pre-trained perception component.

\subsection{ALFRED}\label{sec:alfred}
ALFRED is a benchmark for mapping natural language instructions to actions to accomplish household tasks in the situated environment (\eg~heat an egg). 
Examples in ALFRED come with both single-sentence high-level intents describing a goal (\eg~the NL input in \cref{fig:framework}), and more fine-grained, step-by-step instructions.
In this paper we only use the high-level intents, a more realistic yet more challenging setting to study the effectiveness of our framework in encoding  extra procedural knowledge for under-specified intents.
In addition to the seven atomic actions in the IQA dataset, ALFRED also introduces \actiont{Pickup}, \actiont{Put}, \actiont{ToggleOn}, \actiont{ToggleOff} for object interactions. The argument type of ALFRED is a 2D binary tensor describing regions of the visual space to interact with. Similarly to IQA, we replace the navigation action with an atomic action \aargt{Navigate}{destination}. Previous works also apply similar replacement~\cite{shridhar_alfworld_2020,karamcheti_learning_2020}  %
that allows the agent to proceed to a location without fail. Details in Appendix~\ref{app:experiment}.

\paragraph{Procedure Library}
We create a procedure library for ALFRED by identifying idiomatic control flow and operations from a small set of randomly sampled examples.
The library is designed with two goals in mind as discussed in \S\ref{sec:pap}: \emph{reusability}, where a single function can be applied to multiple similar scenarios, and \emph{coverage}, where a function should cover different execution trajectories under different conditions 
For instance, many tasks consist of a sub-routine to obtain an object by first navigating to the object and then picking up the object by hand, calling for a reusable procedure adaptable to those scenarios (\textbf{C1} in Figure 4). 
Moreover, if an object is positioned inside a receptacle, picking up the object would require opening the receptacle first, an edge case that should be covered by relevant procedures (\eg~(\textbf{C2} in Figure 4)).
Notably, we constrain the conditions of the control flow to the 
logic operation of the property values of objects (\eg~{\tt fridge.is\_openable=True}). 

In total, we define ten such procedural actions, with a complete list in Appendix \ref{app:full_library}.
This creation process is done by the first author, a graduate student proficient in Python. 
The creation of these actions took about two hours, a modest amount of time partially due to PaP's intuitive interface that allows human to summarize complex procedures quickly and partially due to the relative simplicity of the ALFRED, which has a limited number of task types and consistent execution traces.
When we did the sanity check of an initial version of the procedure library (details in Appendix~\ref{sec:alfred_annotation}), we noticed that there were some mismatches.
For example, a laptop should be closed before it is picked up, which was not captured by our library. We thus added a {\tt udp\_close\_if\_needed} function call before the {\tt atomic\_pick\_object} in {\tt udp\_pick\_object}.
On one hand this increases the complexity of the library design process, but on the other hand it also demonstrates the flexibility of the PaP framework, as the necessary fixes could be done entirely by modifying the procedure library itself. \S\ref{sec:iqa_experiment} provides an end-to-end comparison with different procedural libraries.

To investigate the scalability of our annotation process, we also provided a similar guideline and the 21 examples to a separate programmer who does not have any prior knowledge to the dataset. We found that the programmer could quickly understand the PaP Python interface and issue reasonable procedural functions that highly resemble our own creations. This indicates the possibility to curate the procedure libraries with crowd-sourcing efforts. Since our %
More discussion is provided in~\S\ref{sec:discussion} and the full list of the annotation guideline and the user-issued functions are listed in Appendix \ref{sec:user_library}.

\begin{table}[t]
  \centering
  \footnotesize
  \renewcommand{\tabcolsep}{2pt}
  \begin{tabular}{rp{15cm}}
  \toprule
\begin{lstlisting}[numbers=none, basicstyle=\fontfamily{cmtt}\small,style=pythoncode,belowskip=-\baselineskip,aboveskip=- 0.5\baselineskip,commentstyle=\color{codegreen}]
# C1, heat an object with microwave
def udp_heat_object(obj):
    udp_pick_and_put_object(obj, microwave)
    atomic_toggleon_object(microwave)
    atomic_toggleoff_object(microwave)
# C2, prepare the receptacle for future interactions
def udp_prepare_recep(obj):
    reactor = get_reactor("check_obj_attr")
    attr = reactor(obj)
    if attr.is_openable and attr.is_closed:
        atomic_open_object(obj)
\end{lstlisting}
\\ \bottomrule
\end{tabular}
\caption{Two procedural actions for ALFRED %
}
\label{code-prepare}
\end{table}

\paragraph{HMN}~As discussed in \S\ref{sec:hmn}, \textsc{HMN-Planner} generates an executable procedural action $\av^{e}$, given the natural language instruction \xv. We implement our planner with a sequence-to-sequence model with attention \cite{bahdanau2014neural}. Because our implementation is almost identical to the original, we omit the detailed equations.

Based on the construction of the procedure library and the required argument type, we design three reactors: \textsc{AttrChecker}, which has the same functionality as in IQA, \textsc{ReFinder}, which probes where the desired object lies by predicting a receptacle name from all available receptacles to the dataset, and \textsc{MGenerator}, which generates the 2D tensor object masks representing the interaction region. 
Since ALFRED has much richer scene configurations and more diverse objects than IQA, the reactors are fully implemented with neural networks that have strong pattern recognition capabilities. 
This demonstrates the flexibility of our framework to share, add and replace components to suit different situations. 
We describe the detailed implementations of the reactors in \S\ref{sec:alfred_hmn}. The \textsc{HMN} is trained in a supervised fashion, and the heuristic way to induce the supervisions from the original dataset is described in \S\ref{sec:alfred_annotation} in Appendix. 

%% file: 3_experiments.tex
\section{Experiments}
\label{sec:experiments}
We compare our proposed framework with the baseline reactive agents that predicts a single atomic action at each time step. \textcolor{\newcolor}{Notably, we apply the same pretrained vision models, pre-searched map and the \actiont{Navigate} atomic action used in PaP-HMN to the reactive baseline to ensure a fair comparison. More detail in Appendix~\ref{sec:reactive_baseline}}. We attempt to answer the following research questions: (1) Does our framework performs better in complex tasks where there exist inherent hierarchical structures, comparing to a purely reactive system? If so, in what way? (2) Can our framework leverage the procedural knowledge encoded in the procedure library as well as the modularity of its HMN to learn more efficiently? And (3) Can our framework accelerate the development process of the task of interest?

\subsection{Results on IQA}\label{sec:iqa_experiment}
\begin{table}[t]\small
  \centering
  \begin{tabular}{ccccc}
  \toprule
   & & EX & CNT & CT \\
  \midrule
  \multirow{2}{*}{A3C} & \textit{seen} & - & - & - \\
  &\textit{unseen} & 48.6 & 24.5 & 49.9 \\
  \midrule
  \multirow{2}{*}{\textsc{HIMN}} & \textit{seen} & 73.7 & 36.3 & 60.7 \\
  & \textit{unseen} & 68.5 & 30.4 & 58.7 \\
  \midrule
  \multirow{2}{*}{Reactive} & \textit{seen} & 50.0 & 25.1 & 49.6 \\
   & \textit{unseen} & 18.9 & 9.1 & 30.6 \\
  \midrule
  \multirow{2}{*}{PaP-HMN} & \textit{seen} & \textbf{82.8} & \textbf{43.8} & \textbf{82.2} \\
   & \textit{unseen} & \textbf{83.8} & \textbf{45.2} & \textbf{83.1} \\
   \midrule
   PaP$_{v0.1}$-HMN & \textit{seen} & 80.3 & 41.5 & 75.7 \\
  \bottomrule
  \end{tabular}
  \vspace{-0.1cm}
  \caption{The answer accuracy (\%) over IQA dataset on existence (EX), counting (CNT) and contain (CT) questions. The results of AC3 and \textsc{HIMN} are from~\citet{gordon2018iqa}. \textbf{Bold} numbers show the best performance\footnotemark}
  \label{tab:iqa_e2e}
\end{table}

\begin{table}[t]\small
  \centering
  \begin{tabular}{ccc}
  \toprule
   & \textit{seen} & \textit{unseen} \\
  \midrule
  \citet{singh2020moca} & 5.4 & 0.2 \\ 
  Reactive & 21.0 & 5.6\\
  PaP-HMN & \textbf{27.0} & \textbf{11.7} \\
  \midrule
  Reactive + Oracle MG & 40.7 (48.6) & 36.4 (45.0) \\
  PaP-HMN + Oracle MG & \textbf{54.5 (61.0)} & \textbf{51.3 (61.1)} \\
  \bottomrule
  \end{tabular}
  \vspace{-0.1cm}
  \caption{The full task success rate SR (the partial task success rate, SSR, \%) of the baseline reactive model and our model. MG represents the mask generator. Numbers in \textbf{bold} show the best performance for each setting.}
  \label{tab:e2e}
\end{table}

We list the results in \cref{tab:iqa_e2e}\footnotetext{\textit{unseen} features the out-of-distribution visual appearances and arrangements of objects, same for ALFRED}.
Our framework yields the best performance across all models over different question types. 
Through the error analysis, we observe that while the reactive model can generate reasonable action sequences on the \textit{seen} split, its answers are no better than a random guess. This indicates the inability of a reactive model to book-keep the observed objects in the memory.
For \textit{unseen} split, we find that the baseline model could skip some receptacles in its predicted action sequence or even generating syntactically invalid sequences (\eg~functions without required arguments).
This is surprising, since the reactive baseline is trained using the \emph{canonicalized} action sequence according to the roll-out of the for-loops in our procedure library, which is supposed to carry strong patterns (\eg~the target receptacles are ordered from near to far).
This indicates that even simple repeated procedures can be easily represented with a for/while-loop can still be challenging to a reactive agent implemented with a sequence-based backbone. 

\begin{table}[t]
  \centering
  \footnotesize
  \renewcommand{\tabcolsep}{2pt}
  \begin{tabular}{rp{15cm}}
  \toprule
\begin{lstlisting}[numbers=none, basicstyle=\fontfamily{cmtt}\small,style=pythoncode,belowskip=-\baselineskip,aboveskip=- 0.5\baselineskip,commentstyle=\color{codegreen}]
# v0.1: only scan at the start position 
def udp_search_recep():
    r = get_reactor("detect_recep")
    receps = []
    for rotation in range(0, 360, 90):
        atomic_rotate(rotation)
        for horizon in [-30, 0, 30]:
            atomic_look(horizon)
            receps += r()
    return receps
# now: navigate to every reachable point and scan 
def udp_grid_search_recep():
    if not done_search: 
        all_receps = [] # global var
        for pos in reachable_pos:
            atomic_navigate_pos(pos)
            all_receps += udp_search_recep()
    return all_receps
\end{lstlisting}
\\ \bottomrule
\end{tabular}
\caption{Two versions for getting receptacles.}
\label{iqa-diff-library}
\end{table}

\paragraph{Procedure Library Manipulation}~One advantage of our approach is that it decouples the reactors from the creation of the procedural knowledge, thus allowing plug-in update of the procedure library without time-consuming redesigning or retraining the reactors.
\cref{iqa-diff-library} lists two versions of the procedure that decides the list of receptacles to enumerate, and the results of v0.1 are shown at the bottom of \cref{tab:iqa_e2e}. 
In v0.1, the agent stands in its randomly initialized position, looks around, and detects receptacles. Only the detected receptacles are checked to answer the question. 
However, since not all receptacles are visible to the agent at the agent's initial point, such checking could be incomplete. 
We upgraded this procedural function to the new version where the agent searches all possible positions of the scene and memorizes the unmovable receptacle positions. This process only happens once for a scene, and the searched map is stored for future uses. 
In this way, most receptacles are covered. 
This simple modification without changing the remaining parts of the framework improved the CT answer accuracy by 6.6\% as well as improvement of around 2.5\% over the other two question types. 

\subsection{Results on ALFRED}
The ALFRED task success rate (SR) and the sub-task success rate (SSR) are listed in \cref{tab:e2e}. Our model yields a consistent gain over the baseline system on both splits.\footnote{ \citet{singh2020moca} predicts atomic navigation sequences (\eg {\tt MoveAhead}) instead of {\tt Navigate}. The agent struggle to navigate to the target place with only high-level goal. This demonstrates the difficulty of navigation under our experiment setting.}
In our analysis, we find that the Mask R-CNN vision model is the main bottleneck of both end-to-end systems. 
It frequently misclassifies the object types or does not recognize the object in the scene at all. 
This results in the failure of grounding the objects to the environment and thus the failure of the task completion. We hypothesize that the relatively low performance of Mask R-CNN is due to the sub-optimal transfer from the MSCOCO \cite{lin2014microsoft} to the ALFRED data.
Since the development of a better object detector is somewhat orthogonal to our main contributions, to isolate the impact of using a weak object detector on the end-to-end performance, we instead replace the Mask R-CNN with an oracle object mask generator. 
The oracle could always localize and interact with the provided object name if the object is in view for all experiments below. %
We observe a larger performance gap using this oracle mask generator as shown in the bottom half of \cref{tab:e2e}. \textcolor{\newcolor}{This gap suggests that procedural knowledge that could be summarized as several functions describable within a short period of time (in this case, ten functions in two hours) can still be difficult for a reactive system to capture. While the same procedural knowledge can be used in many cases with different environment dynamics, 
a reactive system struggle to distill such knowledge when interacting with highly diverse and dynamic environments.}

\begin{figure}[t]
    \centering
    \includegraphics[width=0.8\columnwidth]{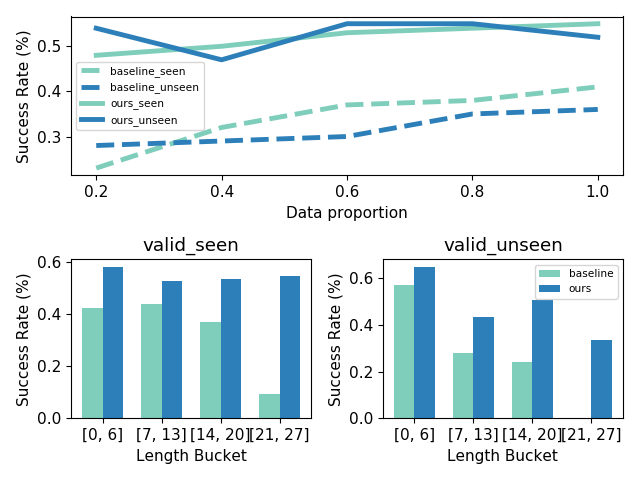}
    \vspace{-0.3cm}
    \caption{The SR (\%) with proportions of the full training set (top) and on each length bucket of the \textit{seen,unseen}~(bottom).}
    \label{fig:data-analysis}
\end{figure}

\paragraph{Performance w.r.t.~Action Length} We further break down the results to different buckets w.r.t~the length of atomic action sequences (without arguments), which roughly represents the difficulty of a task, as shown in \cref{fig:data-analysis}. We observe consistent improvements over all length buckets, %
This difference is even more obvious for challenging tasks with over 21 atomic actions. Our model maintains similar performance for such cases on the \textit{seen} split, and being able to accomplish 30\% tasks successfully on the \textit{unseen} split, while the baseline can barely complete any task. These suggest our framework's stronger capacity to solve long-horizon tasks of deeper hierarchies.

\paragraph{Data Efficiency}
As explained above, our framework uses hierarchical procedural knowledge for planning, which could potentially allow the system to learn complex action sequences in a data-efficient manner.
We benchmark HMN with varying amounts of training data.
As shown in \cref{fig:data-analysis}, with 20\% of the training data, our method exceeds the baseline with the full training set by a large margin (7.7\% and 17.3\% respectively). Furthermore, for the \textit{seen} split, the baseline only obtains less than 60\% SR with 20\% training data, compared to the full data; our method could maintain around 90\% SR of the full data setting. These strongly demonstrate the data efficiency of our method.

\paragraph{Few-shot Generalization}
Next, we test if our framework can generalize to novel compositional procedures with relatively supervised examples. 
We design the few-shot experiments where a subset of the executable procedural actions ($\av^{e}$) are held out, and we sample at most 20 samples of each $\av^{e}$ and add them to the training set. We evaluate the model on these held-out $\av^{e}$. %
We use two strategies to choose the held-out set; the first one is to randomly select $n$ $\av^{e}$; the other is to select the longest $n$ $\av^{e}$ ($n=4$ among 19). PaP-HMN achieves \textbf{33.1} and \textbf{44.9} SR with these two strategies while the reactive baseline only reaches 13.9 and 3.3 respectively\footnote{To reduce the randomness of the \textit{random split}, we report the average of four different splits.}.

Our method consistently outperforms the baseline by a large margin on both settings,
which strongly demonstrates our method's generalization ability in the few-shot scenario.
The significant gain under the short to long setting shows our method's strong capacities in completing long-horizon tasks in a data-efficient way compared to the reactive baseline.

\paragraph{Analysis} Our framework brings several advantages. First, compared to low-level actions, the high-level procedural functions are better aligned with abstract NL inputs. This thus benefits the learning and the prediction of  \textsc{Planner}. Second, programs maintain the consistency of the actions, while a reactive agent might make inconsistent predictions, especially arguments, between actions. Finally, the modular design of the dedicated \textsc{Planner} and the \textsc{Reactors} improve the robust behavior of the agent. More discussion with examples is in Appendix~\ref{sec:advantage}.

Next, we investigate the failures of our framework. First, our ablation study shows that
\textsc{Planner} correctly predicts 80\% executable procedural actions $\av^{e}$, 
and the failures are mainly due to rare words in utterances (\eg~\textit{\underline{soak} a plate}).
In addition, we manually annotated 50 failed examples whose $\av^{e}$ are correct.
We found that 26 failures are due to the sub-optimal interaction positions of the receptacles that we compute during the pre-search phase (\S\ref{sec:pre_search}). 
This causes the interaction with a visible object or receptacle to fail. The pre-search map also missed some objects, and navigating to these objects always failed. 
Additionally, the \textsc{Reactor} prediction errors fail on 18 examples; ambiguous annotations caused two errors, and the wrong argument prediction of the \textsc{Planner} caused four errors. A more comprehensive discussion with the potential solutions is in Appendix~\ref{sec:error_analysis}.

%% file: 5_discussion.tex
\textcolor{\newcolor}{\section{Limitations and Future Work}}\label{sec:discussion}
The experiments demonstrate the benefit of our framework for encoding hierarchical procedural knowledge, especially under low-data or few-shot generalization regimes. However, due to the versatile nature of a complex procedure, the exact conditions and the concrete execution traces are not easy to enumerate. For example, we can heat food using a microwave, an oven or other appliances,  each requiring distinct operations based on the status of that appliance (\eg whether the oven is preheated or not). 
One intuitive solution is to manually create libraries that cover most of the major procedures but fall back to atomic/reactive control when necessary. For example, as in \cref{code-future}, the program calls a reactor implemented as a the neural network ({\tt atomic\_reactor}) to predict atomic actions when using different appliance to heat an object, instead of exhaustively enumerating different conditional branches and their specific action sequences.  
This reactor could be trainable using REINFORCE, or simple maximum likelihood learning if  supervision through action traces is available. 
Another interesting line of research is to automate the procedure library creation through mining and constructing structured procedural knowledge from Web~\cite{tenorth_understanding_2010, kunze2010putting}, or through extracting high-level procedures from corpora of atomic action sequences using unsupervised learning objectives like minimum description length encoding~\cite{ellis2020dreamcoder}.

Though hierarchical procedural knowledge is ubiquitous in human's daily life,
most existing NL instruction following benchmarks do not feature such complex, hierarchical procedures.
Although there can be hierarchies embedded in vision-language navigation tasks~\cite{mattersim}, game playing through reading documentation~\cite{zhong2019rtfm} or through NL communication~\cite{suhr_executing_2019, jernite_craftassist_2019} and mobile phone navigation~\cite{li2020mapping}, the hierarchies are shallow at best, or the occasionally complex ones are limited in their breadth.
This is potentially due to their emphasis on research questions regarding object grounding, spatial relations, interactions and others, and therefore focus less on procedure hierarchies. 
On the other hand, benchmarks that generate examples programmatically (\eg~IQA and ALFRED) often lack realistic and diverse conditional branching in their procedures, as opposed to more free-formed scenarios  discussed above (\cref{code-future}). 
Finally, for those tasks that are more procedurally-complex~\cite{puig_virtualhome:_2018, jernite_craftassist_2019}, experiments are non-trivial due to the lack of automatic evaluation metrics for task completeness. 
Because of this, creating NL instruction following benchmarks that feature more realistic and diverse procedures is one final important direction for future work. 
  
\begin{table}[t]
  \centering
  \footnotesize
  \renewcommand{\tabcolsep}{2pt}
  \begin{tabular}{rp{15cm}}
  \toprule
\begin{lstlisting}[numbers=none, basicstyle=\fontfamily{cmtt}\small,style=pythoncode,belowskip=-\baselineskip,aboveskip=- 0.5\baselineskip,commentstyle=\color{codegreen}]
def udp_heat_object(obj):
    reactor = get_reactor("find_qualified_appliance")
    app = reactor(obj) # (e.g. microwave, oven)
    udp_navigation(app)
    atomic_reactor = get_reactor("predict_atomic_action")
    atomic_action = atomic_reactor(app)
    while atomic_action != STOP:
        env.call(atomic_action)
        atomic_action = atomic_reactor(app)
\end{lstlisting}
\\ \bottomrule
\end{tabular}
\caption{A potential rewriting of \textbf{C1} of \cref{code-prepare}.}
\label{code-future}
\end{table}

\section{Acknowledgement}
We would like to thank Yonatan Bisk for the helpful discussions, Mohit Shridhar for the environment setup and data configurations, Xianyi Cheng, Shruti Rijhwani, Uri Alon, Patrick Fernandes and other Neulab members for feedback on paper, and AWS for donating computational GPU resources used in this work. This work is supported by a grant from the Carnegie Bosch Institute and a contract from the Air Force Research Laboratory under agreement number FA8750-19-2-0200. The U.S. Government is authorized to reproduce and distribute reprints for Governmental purposes notwithstanding any copyright notation thereon. The views and conclusions contained herein are those of the authors and should not be interpreted as necessarily representing the official policies or endorsements, either expressed or implied, of the Air Force Research Laboratory or the U.S. Government.

%% file: 6_appendix.tex
\section{Full Procedural Library}\label{app:full_library}
The full procedural library for IQA is listed in~\cref{code:iqa_full} and that for ALFRED is listed in~\cref{code:alfred_full}.

\section{User-issued Procedural Library}\label{sec:user_library}
\begin{figure*}[t!]
    \centering
    \includegraphics[width=0.7\textwidth]{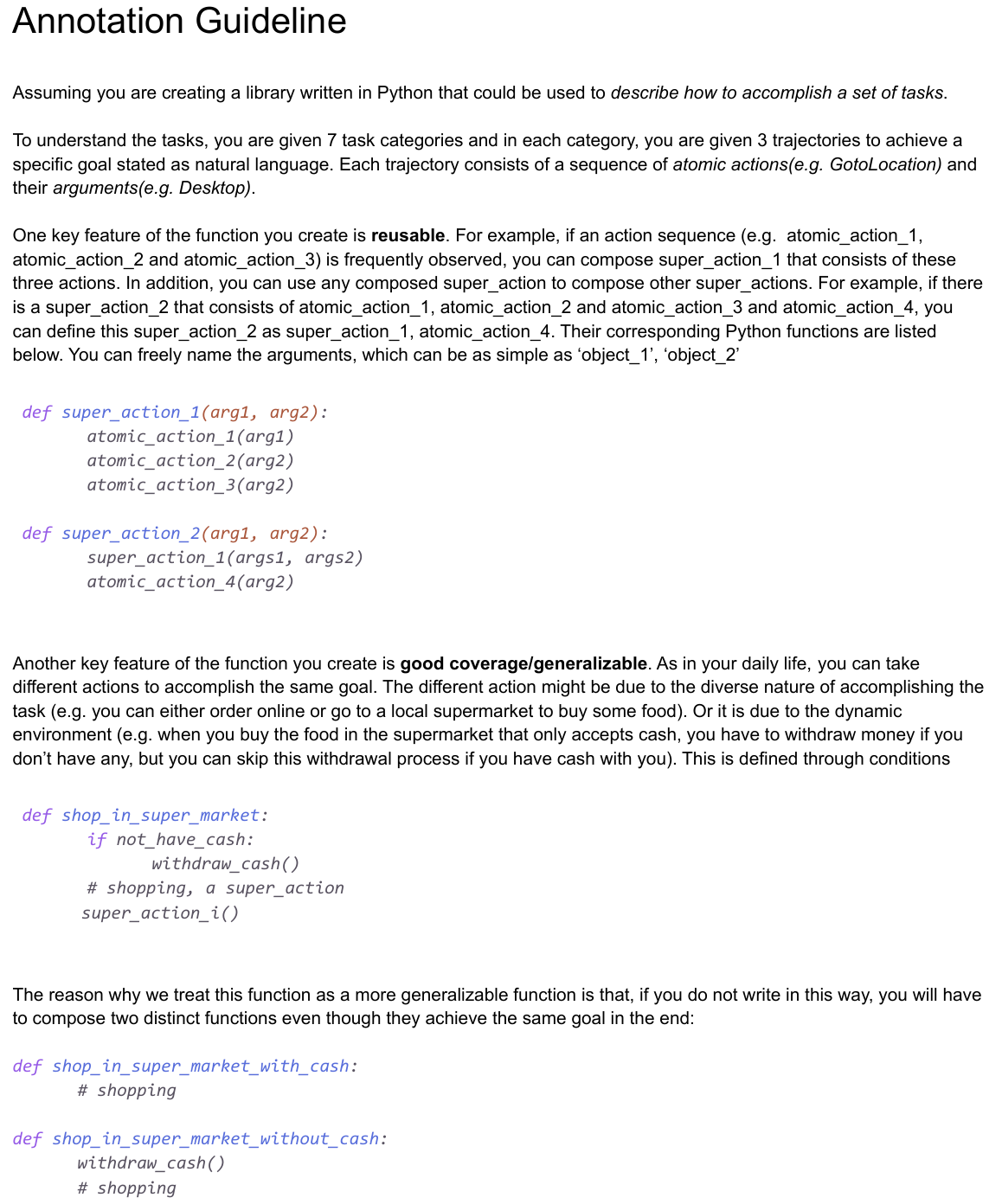}
    \caption{The annotation guideline for a programmer to create procedural functions with 21 examples from the ALFRED dataset.}
    \label{fig:annotation_guide}
\end{figure*}

\cref{fig:annotation_guide} shows the screenshot of the annotation guideline. We purposefully avoid any dataset-related examples. The programmer takes around 90 minutes to complete the annotation. The procedural library created by a programmer without prior knowledge to the ALFRED dataset is in~\cref{code:alfred_human_1}. The programmer could issue reasonable procedural functions that highly resemble our own creations. The reactors can be added to detect the properties of the objects before the condition clauses. 

\section{Experiment Settings}\label{app:experiment}
In this section of the appendix, we describe the detailed implementation of the pre-search map, the heuristic induction of supervisions from existing annotation of the AFLFRED dataset and the implementation of the baseline and our HMN for reproduce purpose. 

\subsection{Pre-search Map Procedure}\label{sec:pre_search}
We treat each scene as a grid map with grid size 0.25. The agent stands on each point, turn around 90 degrees a time and move its camera with degree [-30, 0, 30] and scan. The best position for a receptacle satisfy (1) the agent can open/close the receptacle, can pick up/put an object from/to it. (2) the visual area of the receptacle is the largest compared to other positions. A threshold is used to avoid standing too closed. For ALFRED only, we record the positions of movable objects (\eg apple). This is done by enumerating all the receptacle positions, open them if needed and select the receptacle position that makes the object most visible. 

The map creation also requires an object detection model to detect objects for each scan. For IQA, we use the fine-tuned YOLO-v3 detector as describe in \S\ref{sec:iqa} and the area of an object is calculated by its bounding box. For ALFRED,  we instead use an oracle object detector to minimize the pre-search performance loss. 

Notably, there are many existing works that apply the similar replacement~\cite{shridhar_alfworld_2020,karamcheti_learning_2020}. For example, \citet{shridhar_alfworld_2020} pre-search the map, records the coordinates of each object and uses an A* planner to navigate between two positions. This replacement 
that allows the agent to proceed to a location without fail.

\subsection{Reactive Baseline}\label{sec:reactive_baseline}
\paragraph{IQA}~The reactive baseline is implemented as a pointer network \cite{vinyals2015pointer} whose output sequence corresponds to the positions in an input sequence. To make a fair comparison with our method, we provide this baseline with the available receptacle IDs of each scene, the question type, and the targeted objects. For instance, given the question \textit{how many mugs in the fridge} for scene $i$, we list all the receptacles (\eg fridge\_1, cabinet\_2) in the order of distance to the agent's initial position as well as the question type ``contains'' and the two working objects ``mug'' and ``fridge''. The fixed set of actions and the answers are added at the beginning of the input so that the model does not need an extra generation component. The reactive agent needs to navigate to each receptacle, operate them properly and generate an answer at the end. The images are encoded and the objects are detected with the same YOLO-v3 detector as in HMN.

While an action sequence is not provided in the release of the dataset, we heuristically create such action sequences by enumerating the input receptacle list of each sample. The size for each question type is 7000 and a total of 21000 samples are used in the training. We additionally compare with the \textsc{HIMN} proposed in \citet{gordon2018iqa} that designs a meta-controller that calls different controllers to accomplish different tasks (\eg navigation, manipulation), and an A3C agent implemented in the same work. 

\paragraph{ALFRED}~We follow \citet{shridhar_alfred:_2019} to setup our reactive baseline. This baseline takes the natural language instruction \xv as input, then it predicts an atomic action at each time step, conditioned on the vision, the previous generated atomic action, and the attended language. The baseline also has a progress monitor component to track the task completion progress~\cite{ma_self-monitoring_2019}. We make the same replacement of the atomic navigation actions with \aargt{Navigate}{destination}. The original mask generator is replaced by the same Mask R-CNN used in our HMN. 

For both datasets, we use \texttt{seen} and \texttt{unseen} validation set for the evaluation. The floorplans 
of the \texttt{unseen} split are held-out in the training data. Each floorplan defines the appearance of the environment as well as the arrangement of the objects. For IQA, we measure the answer accuracy, and we follow \citet{shridhar_alfred:_2019} to measure the task success rate (SR), which defines the percentage of whole task completion; and sub-task success rate (SSR), which measures the ratio of individual sub-task completion for ALFRED.

\subsection{HMN Implementation}
\label{sec:iqa_hmn}\label{sec:alfred_hmn}
\paragraph{IQA} Since the natural language questions \xv are generated with a limited number of templates, we use a rule-based \textsc{HMN-Planner} that recognizes each template and classifies a template to one of the three question types whose corresponding procedural actions are listed as the top three functions in \cref{code:iqa_full}. 

We model the two reactors \textsc{AttrChecker} and \textsc{RelChecker} as two multi-classes classifiers. We first follow \citet{gordon2018iqa} to use a YOLO-v3 \cite{redmon2018yolov3} that is fine-tuned on the images sampled from THOR for object detection. This object detector scan each visual input and generate a bounding box and a class name for each detected object. Since there are only seven receptacles, the \textsc{AttrChecker} uses the predicted class name of a receptacle to decide whether the receptacle is openable or not. It then marks the receptacle as is\_open=True after the atomic open action is launched for the receptacle. The \textsc{RelChecker} use bounding box to heuristically decide the spatial relation between an object and a receptacle. The \textsc{RelChecker} considers that an object is inside a receptacle if its bounding box has over 70\% overlap with the receptacle's bounding box.

\paragraph{ALFRED} We use a sequence-to-sequence model with attention~\cite{bahdanau2014neural} as our \textsc{Planner}. The input to the encoder is the natural language \xv. The decoder generates one function $a_i$ at a time from a constrained vocabulary $\Ap \bigcup \Aa$, conditioned on \xv and the action history $\{a_1, ..., a_{i-1}\}$.

We adopt the pre-trained Mask R-CNN \cite{he2017mask} that is fine-tuned on the ALFRED dataset from \citet{shridhar_alfworld_2020} as our \textsc{MGenerator}. It returns the name and the bounding box for all detected objects in the visual input. Its parameters are frozen. We design \textsc{AttrChecker} and \textsc{ReFinder} as two multi-classes classifiers.
The inputs to these two reactors are the object name $\bm{h}^o$ encoded by a \textsc{Bi-LSTM}, the immediate vision $\bm{h}^i$ encoded by a frozen \textsc{ResNet-18} CNN \cite{he2016deep} following \citet{shridhar_alfred:_2019}, the called action sequence $\bm{h}^a$ encoded with a \textsc{LSTM} and the attended input $\bm{h}^l$ with $\bm{h}^a$. These four vectors are concatenated together as $\bm{h}^f$. A fully connected layer and a non-linear activation function are added to predict class probabilities. 

\subsection{AFLRED Supervision Induction}
\label{sec:alfred_annotation}
We induced the ground truth labels for each component of the HMN from ALFRED with the help of atomic action sequences and the subgoal sequences provided by the dataset so that the HMN can be trained in a supervised fashion to maximize the log-likelihood of the label. First, we used the subgoal sequences to annotate the executable procedural actions for the planner. For example, a subgoal sequence \actiont{Goto}, \actiont{Pick}, 
\actiont{Clean}, \actiont{Goto}, \actiont{Put} was annotated with {\tt udp\_clean\_object}, {\tt udp\_put\_object}. A different subgoal sequence \actiont{Goto}, \actiont{Pickup}, \actiont{Goto}
\actiont{Clean}, \actiont{Put} was annotated with the same procedural action sequence. The first author annotated 30 most frequent subgoal sequences of the training set of ALFRED and resulted in 19 different executable procedural actions\footnote{We discarded a training example if its subgoal sequence is not annotated with the procedure library. About 500 samples among 21k training data are discarded.}. Next, we used the atomic action sequences of the dataset to generate the labels for the reactors. For example, if there is an \actiont{Open} before a \actiont{Pickup} in the atomic action sequence, the attribute of the corresponding object is labeled as openable=True and is\_open=False.

When doing the sanity check to verify the coverage of our created procedural library, we assign an executable procedural action $\av^{e}$ to each sample, we then check whether the atomic action sequence of $\av^{e}$ match the annotated atomic action sequence provided by the dataset. Unmatched examples are reviewed and the procedural library is updated as in \S\ref{sec:alfred}.

\subsection{Hyperparameters}
\paragraph{IQA Baseline} The embedding size is 100, the hidden size of the \textsc{Bi-LSTM} and \textsc{LSTM} are 256 and 512. We take the same three feature vectors before the YOLO detection layer and convert the channel size to 32 with convolution layers to encode an image. The flatted features are concatenated with dropout rate of 0.5. We use Adam~\cite{kingma2014adam} with learning rate 1e-4. 

\paragraph{ALFRED} We follow \citet{shridhar_alfred:_2019} for the hyperparameter selection of the baseline and our model if they are applicable (\eg embedding size, optimizer). We observe that training longer yields better task completion, and thus we train the baseline for 15 epochs and ours for 10 epochs. For our method only, the size of $\bm{h^o}$, $\bm{h^a}$ and $\bm{h^l}$ is 512. The activation function of \textsc{AttrChecker} is Sigmoid and the output size is 3 (\ie is\_openable, is\_open, is\_close). The activation function of \textsc{ReFinder} is Softmax and the output size equals the object vocabulary size. 

\begin{table*}[t]\scriptsize
    \centering
    \begin{tabular}{l}
    \toprule
    \textbf{Task:}~Put a chilled egg in the sink\\
    \textbf{Reactive: }~\aargt{Navigate}{egg} \aargt{Pickup}{egg} \aargt{Navigate}{fridge} \aargt{Open}{fridge} \textcolor{blue}{{\tt STOP}}\\
    \textsc{\textbf{HMN-Planner:}}~{\tt udp\_cool\_object}({\tt egg}), \tt{udp\_pick\_put\_object}({\tt egg}, {\tt sink})\\
    \midrule
    \textbf{Task:}~Put CDs in a safe. (*requires to put \emph{two} CDs)\\
    \textbf{Reactive: }~\aargt{Navigate}{cd} \aargt{Pickup}{cd} \aargt{Navigate}{safe} \aargt{Open}{safe} \aarg{Put}{cd}{safe} \aargt{Close}{safe} \textcolor{blue}{{\tt STOP}}\\
    \textsc{\textbf{HMN-Planner:}}~{\tt udp\_pick\_put\_object}({\tt cd}, {\tt safe}), {\tt udp\_pick\_put\_object}({\tt cd}, {\tt safe})\\
    \midrule
    \textbf{Task:}~Place a cooked potato slice in the fridge\\
    \textbf{Reactive:}~\aargt{Navigate}{knife} \aargt{Pickup}{knife} \aargt{Navigate}{potato} \aargt{Slice}{potato}  \aargt{Navigate}{\underline{fridge}} \\ \aarg{Put}{knife}{\underline{countertop}} \aargt{Navigate}{potato}{} \aargt{Close}{potato} ...\\
    \textsc{\textbf{HMN-Planner:}}~{\tt udp\_slice\_object}({\tt potato}), {\tt udp\_heat\_object}({\tt potato}), {\tt udp\_pick\_and\_put}({\tt potato}, {\tt fridge})\\
    \bottomrule
    \end{tabular}
    \vspace{-0.2cm}
    \caption{Common failures of the reactive baseline. All actions of the reactive baseline are atomic actions.}
    \label{tab:baseline_fail}
\end{table*}

\section{Analysis}
In this section, we present concrete examples to demonstrate the benefit of our proposed pipeline. We also show a few failures of our pipeline to encourage future developments.
\subsection{Advantage of HMN}
\label{sec:advantage}
The above results suggest that our proposed framework with modularized task-specific components and predefined procedure knowledge is effective in controlling situated agents via complex natural language commands.
Compared with the reactive agent, this framework brings several benefits. 
First, instead of directly controlling an agent using low-level atomic actions, it predicts holistic procedural programs, which are better aligned with high-level input NL descriptions.
For instance, in Examples 1 and 2 in \cref{tab:baseline_fail}, common NL phrases like \textit{put $\cdot$ in $\cdot$} naturally map to the procedure {\tt udp\_pick\_put\_object}, while the  reactive baseline could struggle at interpreting the correspondence between the NL intents and the verbose low-level atomic actions, resulting in incomplete predictions.
Second, using procedures could help maintain \emph{consistency} of actions.
Specifically, given a procedure (\eg {\tt udp\_pick\_put\_object}), and its arguments (\eg {\tt knife}, {\tt fridge}), the HMN agent is guaranteed to coherently carry out the specified action without being interfered, while the reactive baseline could predict inconsistent atomic actions in-between (\eg~the underscored arguments of {\tt Navigate} and {\tt Put} should be the same in Example 3).
Finally, we remark that procedures also improve the robust behavior of the agent.
For instance, when interacting with container objects (\eg~{\tt fridge}), HMN would call the dedicated \textsc{AttrChecker} to decide whether to open the object first  (\eg~\texttt{C4},Fig. 1), and it mis-predicts once, while the reactive baseline fails to perform the \actiont{Open} action 33 times on the \textit{unseen} split.

\subsection{Error Analysis}
\label{sec:error_analysis}
We first did an ablation study on the \textsc{Planner} on the \textit{unseen} split.
\textsc{Planner} correctly predicts 80\% executable procedural actions $\av^{e}$, 
and the failures are mainly due to rare words in utterances (\eg~\textit{\underline{soak} a plate}.
Next, we manually annotated 50 failed examples among samples whose $\av^{e}$ are correctly predicted by the \textsc{Planner}.
We found that 26 failures are due to the sub-optimal interaction positions of the receptacles that we compute during the pre-search phase (\S\ref{sec:pre_search}). 
This results in the failures of putting an object in-hand to a visible receptacle or picking up a visible object. The pre-search map also missed some objects and navigating to these objects always failed. This problem can be alleviated either by adding additional procedural actions to move around and attempt to pick up or put an object until success, or by doing more careful engineering to create the map. 
Additionally, 18 examples are caused by prediction errors of reactors.
For instance, \textsc{ReFinder} could given incorrect predictions of the containing receptacle of an object. The receptacle is not correctly operated before the targeted object is visible.
While such errors are inevitable due to imperfect reactors, it could be potentially mitigated by designing more robust procedures, \eg, enumerating over the top-$n$ most likely receptacles for a target object instead of the best scored one by the reactor.
Other approaches, like introducing object-centric representations to the reactors~\citep{wu2017learning, singh2020moca}, could also be helpful.
The remainder of the errors are caused by ambiguous annotation (2 examples), and wrong argument predictions of the planner (4 examples).

\input{4_related_works}

\input{full_library}

%% file: 4_related_works.tex
\section{Related Work}
\label{sec:related_work}
\paragraph{Procedure-guided Learning}
The idea of using predefined procedures for agent control has been explored in the literature.
For example, \citet{andreas_modular_2017, das_neural_2019} use high-level symbolic program sketches to guide an agent's exploration; \citet{gordon2018iqa, yu2019multi} design meta-controller to call different low-level controllers. There only exists one explicit level of the hierarchy.
\citet{sun_program_2020} show that programs can assist agent's task completions. They require the presence of the program for each task, while our programs are generated by the planner. There is no nested function in their provided programs too.
Programs are used to represent procedures in \citet{puig_virtualhome:_2018}, but no hierarchy is considered. Later \citet{liao_synthesizing_nodate} annotate the dataset with program sketches and propose a graph-based method to generate executable programs. Their work requires a fully observed environment while we only consider egocentric visions.
Recent works also explore representing hierarchies with natural language~\cite{hu2019hierarchical,jiang_language_2019} and visual goal representation~\cite{misra2018mapping} instead of symbols. 
Another related area is probabilistic programming, where procedures serve as symbolic scaffolds to define the control flow of learnable programs \citep{gaunt2017differentiable}.
Our work is related to these research in using predefined procedural knowledge to assist learning, while we focus on leveraging such procedures to synthesize executable programs from natural language commands.

\paragraph{Semantic Parsing}~Our work is also related to semantic parsing, where executable programs are generated from natural language inputs. 
This includes mapping NL to domain-specific logical forms (\eg~lambda calculus, \cite{Zettlemoyer2005LearningTM}) or programs (\eg~SQL, \cite{Zhong2017Seq2SQLGS,yu2018spider}).
Recently there has also been a burgeoning of developing models that could transduce natural language intents into general-purpose programs (\eg~Python, \cite{yin_syntactic_2017,Rabinovich2017AbstractSN}).
Our work also considers program generation from NL, with a focus on the command and control of situated agents.

Research in semantic parsing has also explored leveraging idiomatic program structures, which are fragments of programs that frequently appear in training data, to aid generation~\citep{Raghothaman2016SWIMSW}. 
Such idiomatic programs are mined from corpora~\citep{iyer-etal-2019-learning,shin_program_2019}.
Our work focuses on designing flexible and idiomatic procedures which interact with situated components (\eg~reactors) to adapt to environment-specific situations. 
This work also uses manually-curated procedures, because in our problem setting we do not have a readily available corpus of high-level procedural programs to automatically collect such idioms.
We leave extracting procedures from low-level atomic actions as interesting future work.

\paragraph{Robotics Planning and Hierarchical Control} Our procedure library shares the design philosophy with the macro-actions in the STRIPS representation in the robotics planning \cite{fikes1971strips}. However, we do not define the pre-condition and the post-effect of the actions, and instead leave the models to learn the consequences. 
The task-level planning has been studied extensively~\cite{kaelbling2011hierarchical,srivastava2013using, srivastava2014combined}. These methods often work with high-level formal languages in low-dimensional state space, and they are typically designed for a specific environment and task. Our framework can be applied to various tasks and only partial observations are required. Previous works also leverage PDDL and the answer set planner (ASP) for task planning. PDDL+ASP is conceptually different from our formalism. PDDL+ASP aims at planning the actual execution sequences. The PDDL planner searches the action sequences based on the initial and the final state. Meanwhile, our formalism focuses on describing the procedure to accomplish a task. We use the HMN-Planner to predict the executable procedure sequence given the NL. It is possible to integrate them into one system. E.g.,a procedure function can call a PDDL planner if the pre/post conditions are clearer given NL. 
Finally, many works design mechanism to learn hierarchies automatically from supervisions of only the end-task~\cite{sutton1999between, bacon2017option}, which might suffer from collapsing to trivial atomic actions.

%% file: full_library.tex
\begin{table}[htbp]
  \centering
  \footnotesize
  \renewcommand{\tabcolsep}{2pt}
  \begin{tabular}{rp{15cm}}
  \toprule
\begin{lstlisting}[numbers=none, basicstyle=\fontfamily{cmtt}\small,style=pythoncode,belowskip=-\baselineskip,aboveskip=- 0.5\baselineskip,commentstyle=\color{codegreen}]
# check the existence of an object in the scene
def udp_check_obj_exist(obj):
    all_recep = udp_grid_search_recep()
    for recep in all_recep:
        rel = udp_check_relation(obj, recep)
        if rel == OBJ_IN_RECEP:
            return True
    return False

# check whether a receptacle contains an object
def udp_check_contain(obj, recep):
    all_recep = \
    udp_grid_search_tar_recep(recep.desc)
    for recep in all_recep:
        rel = udp_check_relation(obj, recep)
        if rel == OBJ_IN_RECEP:
            return True
    return False

# count how many objects in the scene  
def udp_count_obj(obj):
    tot = 0
    all_recep = udp_grid_search_recep()
    for recep in all_recep:
        rel = udp_check_relation(obj, recep)
        tot += int(rel == OBJ_IN_RECEP)
    return tot

# check object inside receptacle
def udp_check_relation(obj, recep):
    atomic_navigate(recep)
    r1 = get_reactor("check_obj_attr")
    r2 = get_reactor("check_obj_recep_rel")
    attr = r1(recep)
    if attr.is_openable and attr.is_closed:
        atomic_open_object(recep)
        rel = r2(obj, recep)
        atomic_close_object(recep)
    else:
        rel = r2(obj, recep)
    return rel
    
# get a list of target receptacles
def udp_grid_search_tar_recep(desc):
    recep_list = udp_grid_search_recep()
    tar_recep_list = [x for x in recep_list \\
    if x.desc == desc]
    return tar_recep_list
# navigate and search at every reachable points  
def udp_grid_search_recep():
    if not done_search:
        all_receps = [] # global var
        for pos in reachable_pos:
            atomic_navigate_pos(pos)
            all_receps += udp_search_recep()
    return all_receps
\end{lstlisting}
\\ \bottomrule
\end{tabular}
\caption{Procedural functions defined for IQA}
\label{code:iqa_full}
\end{table}

\begin{table}[htbp]
  \centering
  \footnotesize
  \renewcommand{\tabcolsep}{2pt}
  \begin{tabular}{rp{15cm}}
  \toprule
\begin{lstlisting}[numbers=none, basicstyle=\fontfamily{cmtt}\small,style=pythoncode,belowskip=-\baselineskip,aboveskip=- 0.5\baselineskip,commentstyle=\color{codegreen}]
# close an object if it is open
def udp_close_if_needed(obj):
    reactor = get_reactor("check_obj_attr")
    attr = reactor(obj)
    if attr.is_openable and attr.is_open:
        atomic_close_object(obj)
# "postpare" the receptacle
def udp_postpare_recep(obj):
    reactor = get_reactor("check_obj_attr")
    attr = reactor(obj)
    if attr.is_openable and attr.is_open:
        atomic_close_object(obj)
# pickup an object       
def udp_pick_object(obj):
    reactor = get_reactor("find_obj_recep")
    udp_navigation(obj)
    recep = reactor(obj)
    udp_prepare_recep(recep)
    # this is for pickup laptop only
    udp_close_if_needed(obj)
    atomic_pick_object(obj)
    udp_postpare_recep(recep
# put an object to a receptacle
def udp_put_object(obj, dst):
    udp_navigation(dst)
    udp_prepare_recep(dst)
    atomic_put_object(obj, dst)
    udp_postpare_recep(dst)
# clean an object in the fauucet    
def udp_clean_object(obj):
    # sink and faucet are global variables
    udp_pick_object(obj)
    udp_put_object(obj, sink)
    atomic_toggleon_object(faucet)
    atomic_toggleoff_object(faucet)
    udp_pick_object(obj)
# slice an object with a knife
def udp_slice_object(obj, tool_dst):
    # knife is a global variable
    udp_pick_object(knife)
    udp_navigation(obj)
    reactor = get_reactor("find_obj_recep")
    recep = reactor(obj)
    udp_prepare_recep(recep)
    atomic_slice_object(obj)
    udp_postpare_recep(recep)
    udp_put_object(tool, tool_dst)
# pick an object and then put it to a receptacle   
def udp_pick_and_put_object(obj, dst):
    udp_pick_object(obj)
    udp_put_object(obj, dst)
# cool an object with fridge 
def udp_cool_object(obj):
    # fridge is a global variable
    udp_pick_and_put_object(obj, fridge)
# heat an object with microwave
def udp_heat_object(obj):
    udp_pick_and_put_object(obj, microwave)
    atomic_toggleon_object(microwave)
# prepare a receptacle for interaction
def udp_prepare_recep(obj):
    reactor = get_reactor("check_obj_attr")
    attr = reactor(obj)
    if attr.is_openable and attr.is_closed:
        atomic_open_object(obj)
\end{lstlisting} 
\\ \bottomrule
\end{tabular}
\caption{Procedural functions defined for ALFRED}
\label{code:alfred_full}
\end{table}

\begin{table}[htbp]
  \centering
  \footnotesize
  \renewcommand{\tabcolsep}{2pt}
  \begin{tabular}{rp{15cm}}
  \toprule
\begin{lstlisting}[numbers=none, basicstyle=\fontfamily{cmtt}\small,style=pythoncode,belowskip=-\baselineskip,aboveskip=- 0.5\baselineskip,commentstyle=\color{codegreen}]
# udp_pick_object(obj):
def udp_pick_up(object, loc):
    udp_navigation(loc)
    if loc.is_open:
        atomic_pickup_object(object)
    else:
        atomic_open_object(loc)
        atomic_pickup_object(object)
        atomic_close_object(loc)

def udp_pick_up_to(object, loc, loc_to):
    udp_pick_up(object, loc)
    udp_navigation(loc_to)

# udp_put_object(obj, dst):
def udp_put_to(object, loc_to):
    udp_navigation(loc_to)
    if loc.is_open:
        PutObject(object)
    else:
        atomic_open_object(loc_to)
        atomic_put_object(loc_to)
        atomic_close_object(loc_to)

# udp_pick_and_put_object(obj, dst):
def udp_pick_put_to(object, loc, storage):
    udp_pick_up(object, loc)
    udp_put_to(object, storage)

def udp_look_under_light(object, loc, light_source):
    udp_pick_up_to(object, loc, light_source)
    atomic_toggleon_object(light_source)
    
# udp_slice_object(obj, tool_dst):
def udp_slice(object, loc, slicer):
    udp_pick_up_to(slicer, loc, object)
    atomic_slice_object(object)

def udp_toggle(object):
    atomic_toggleon_object(object)
    atomic_toggleoff_object(object)

# udp_cool_object(obj):
def udp_cool(object, loc):
    udp_pick_put_to(object, loc, fridge)

# udp_heat_object(obj):
def udp_heat(object, loc):
    udp_pick_put_to(object, loc, microwave)
    udp_toggle(microwave)
    udp_pick_up(object, microwave)

# udp_clean_object(obj):
def udp_clean(object, loc):
    udp_pick_put_to(object, loc, Faucet)
    udp_toggle(Faucet)
\end{lstlisting} 
\\ \bottomrule
\end{tabular}
\caption{Procedural functions defined by a programmer without ALFRED domain knowledge. The comments could roughly map to functions in \cref{code:alfred_full}.}
\label{code:alfred_human_1}
\end{table}    